%% file: acl_latex.tex
\pdfoutput=1
\documentclass[11pt]{article}

\usepackage[preprint]{acl}
\usepackage{markdown}
\usepackage{times}
\usepackage{latexsym}

\usepackage[T1]{fontenc}
\usepackage[utf8]{inputenc}

\usepackage{microtype}

\usepackage{inconsolata}

\usepackage{graphicx}

\usepackage{algorithm}
\usepackage[algo2e,linesnumbered,ruled,vlined]{algorithm2e}
\usepackage{algorithmic}

\usepackage{tcolorbox}
\usepackage{seqsplit}
\usepackage{makecell}
\usepackage{textcomp}
\tcbuselibrary{theorems,breakable,skins}
\newtcbtheorem[]{myexample}{Prompt}%
{
  breakable,
  enhanced,
  colback=green!5,
  colframe=green!35!black,
  fonttitle=\bfseries,
  top=2pt,         % Space between top frame and content
  bottom=2pt,      % Space between bottom frame and content
  before skip=4pt, % Space before the box
  after skip=4pt   % Space after the box
}{th}

\usepackage{hyperref}       % hyperlinks
\usepackage{url}            % simple URL typesetting
\usepackage{booktabs}       % professional-quality tables
\usepackage{amsfonts}       % blackboard math symbols
\usepackage{nicefrac}       % compact symbols for 1/2, etc.
\usepackage{microtype}      % microtypography
\usepackage{lipsum}
\usepackage{fancyhdr}       % header
\usepackage{graphicx}       % graphics
\usepackage{float}
\usepackage{multirow}
\usepackage[utf8]{inputenc}
\graphicspath{{media/}}     % organize your images and other figures under media/ folder
\usepackage{siunitx} %正确显示浓度值并包括正确的单位
\usepackage{gensymb} % 提供了 \degree 命令
\usepackage{markdown}
\usepackage{tabularx}
\usepackage{xcolor}
\usepackage{hyperref}
\usepackage{multirow}
\usepackage{listings}
\usepackage{longtable}
\usepackage{bbding}
\usepackage{enumitem}
\usepackage{amssymb}
\usepackage{wrapfig}
\usepackage{array}
\hypersetup{
    colorlinks=true, % 设置为true表示链接带颜色,设为false则表示链接以框的形式呈现
    linkcolor=blue, % 内部链接（例如目录到章节的链接）的颜色设置为蓝色
    filecolor=magenta, % 文件链接的颜色
    urlcolor=cyan, % 外部URL链接的颜色
    citecolor=green % 引用链接的颜色
}

\newtcolorbox{prompt}[1]{
    enhanced,
    colback=gray!20,
    colframe=black,
    boxrule=0.3pt,
    arc=3mm,
    left=2pt,
    right=2pt,
    boxsep=3pt,
    fonttitle=\small\bfseries,
    title=#1,
    fontupper=\scriptsize
}

\definecolor{lightgray}{gray}{0.9}
\DeclareUnicodeCharacter{FF08}{\textlparen} % Unicode for ‘（’
\DeclareUnicodeCharacter{FF09}{\textrparen} % Unicode for ‘）’

\title{SA-GCS: Semantic-Aware Gaussian Curriculum Scheduling for UAV Vision-Language Navigation}

\author{%
\parbox{\linewidth}{\centering%
\textbf{Author Name$^{1}$\thanks{Equal Contribution}} and \textbf{Author Name$^{1,2}$} \\[2ex] 
$^1$Institution Name \quad $^2$Institution Name \\[2ex] 
$^3$Institution Name \quad $^4$Institution Name \\[2ex] 
$^5$Institution Name \quad $^6$Institution Name \\[2ex] 
$^7$Institution Name \\[2ex] 
}%
}

\author{%
\parbox{\linewidth}{\centering%
\textbf{Hengxing Cai$^{1,2}$\thanks{Equal Contribution}}, 
\textbf{Jinhan Dong$^{2,3}$\footnotemark[1]}, 
\textbf{Yijie Rao$^{2,4}$\footnotemark[1]}, 
\textbf{Jingcheng Deng$^5$},
\textbf{Jingjun Tan$^{1}$}, \\[0.5ex]
\textbf{Qien Chen$^{1}$},
\textbf{Haidong Wang$^1$},  
\textbf{Zhen Wang$^2$}, 
\textbf{Shiyu Huang$^6$}, 
\textbf{Agachai Sumalee$^7$} \\[0.5ex] 
and \textbf{Renxin Zhong$^{1}$\thanks{Corresponding author}} \\[2ex]
$^1$School of Intelligent Systems Engineering, Sun Yat-Sen University \quad 
$^2$DP Technology \quad 
$^3$Beijing University Of Posts and Telecommunications \quad 
$^4$Beihang University \\ 
$^5$Institute of Computing Technology, Chinese Academy of Sciences \quad 
$^6$XPENG \\ 
$^7$School of Integrated Innovation, Chulalongkorn University \\[5ex]
}%
}

\begin{document}
\maketitle
\input{chapters/0_abs}

\input{chapters/1_intro}
\input{chapters/2_related_work}

\input{chapters/3_method}
\input{chapters/4_experiments}

\input{chapters/5_results_and_analysis}
\input{chapters/6_conclusion}

\clearpage
\bibliography{custom}

\clearpage
\appendix
\input{chapters/99_appendix}

\end{document}

%% file: chapters/0_abs.tex
\begin{abstract}
Unmanned Aerial Vehicle (UAV) Vision-Language Navigation (VLN) aims to enable agents to accurately localize targets and plan flight paths in complex environments based on natural language instructions, with broad applications in intelligent inspection, disaster rescue, and urban monitoring. 
Recent progress in Vision-Language Models (VLMs) has provided strong semantic understanding for this task, while reinforcement learning (RL) has emerged as a promising post-training strategy to further improve generalization. 
However, existing RL methods often suffer from inefficient use of training data, slow convergence, and insufficient consideration of the difficulty variation among training samples, which limits further performance improvement.
To address these challenges, we propose \textbf{Semantic-Aware Gaussian Curriculum Scheduling (SA-GCS)}, a novel training framework that systematically integrates Curriculum Learning (CL) into RL. 
SA-GCS employs a Semantic-Aware Difficulty Estimator (SA-DE) to quantify the complexity of training samples and a Gaussian Curriculum Scheduler (GCS) to dynamically adjust the sampling distribution, enabling a smooth progression from easy to challenging tasks. 
This design significantly improves training efficiency, accelerates convergence, and enhances overall model performance.
Extensive experiments on the CityNav benchmark demonstrate that SA-GCS consistently outperforms strong baselines across all metrics, achieves faster and more stable convergence, and generalizes well across models of different scales, highlighting its robustness and scalability. 
The implementation of our approach is included in the supplementary materials and will be made publicly available in the future.
\end{abstract}

%% file: chapters/1_intro.tex
\section{Introduction}

Vision-Language Navigation (VLN) for Unmanned Aerial Vehicles (UAVs) is an emerging research area that focuses on enabling UAVs to interpret natural language instructions for target localization and navigation in complex environments.
This task holds significant potential in applications such as intelligent inspection, aerial search and rescue, and urban surveillance. To improve performance on UAV VLN tasks, early studies have explored various approaches, including supervised sequence-to-sequence (Seq2Seq)~\cite{anderson2018vision}, Cross-Modal Attention (CMA)~\cite{liu2023aerialvln}, and Map-based Goal Prediction (MGP)~\cite{lee2024citynav}. These methods have significantly improved the model's capabilities in visual target recognition, natural language instruction understanding, and path reasoning for navigation.

More recently, the integration of Large Language Models (LLMs), especially Vision-Language Models (VLMs), into UAV VLN has shown remarkable progress. These models offer strong semantic alignment and understanding capabilities, which improve systems' capability to parse complex instructions and perceive environments, thereby leading to better navigation decisions~\cite{li2025skyvln, zhang2025logisticsvln, ping2025multimodal}. Building on this, some recent works have further explored post-training strategies tailored for the UAV VLN~\cite{saxena2025uav, cai2025flightgpt}. Among them, Reinforcement Learning (RL), which optimizes policies through interaction with the environment and reward signals, has shown stronger potential than Supervised Fine-Tuning (SFT) in continuously adapting to dynamic environments and improving model generalization~\cite{cai2025flightgpt}.

Despite the benefits of RL in UAV VLN, the training process faces several critical challenges. 
On the one hand, policy learning is often hindered by instability, low sample efficiency, and slow convergence.
% On the other hand, and more importantly, training samples vary significantly in difficulty. 
% Direct use of a mixed-difficulty dataset tends to result in overfitting on simpler examples while under-training on harder ones, thereby limiting overall model performance.
% On the other hand, training samples vary significantly in difficulty, and directly using a mixed-difficulty dataset often leads to overfitting on simpler examples while leaving harder ones under-trained, thereby limiting overall model performance.
On the other hand, training samples vary significantly in difficulty, with the direct use of a mixed-difficulty dataset often leading to overfitting on simpler examples while leaving harder ones under-trained, thereby limiting overall model performance.

\begin{figure}[t]
\centering
\includegraphics[width=0.47\textwidth]{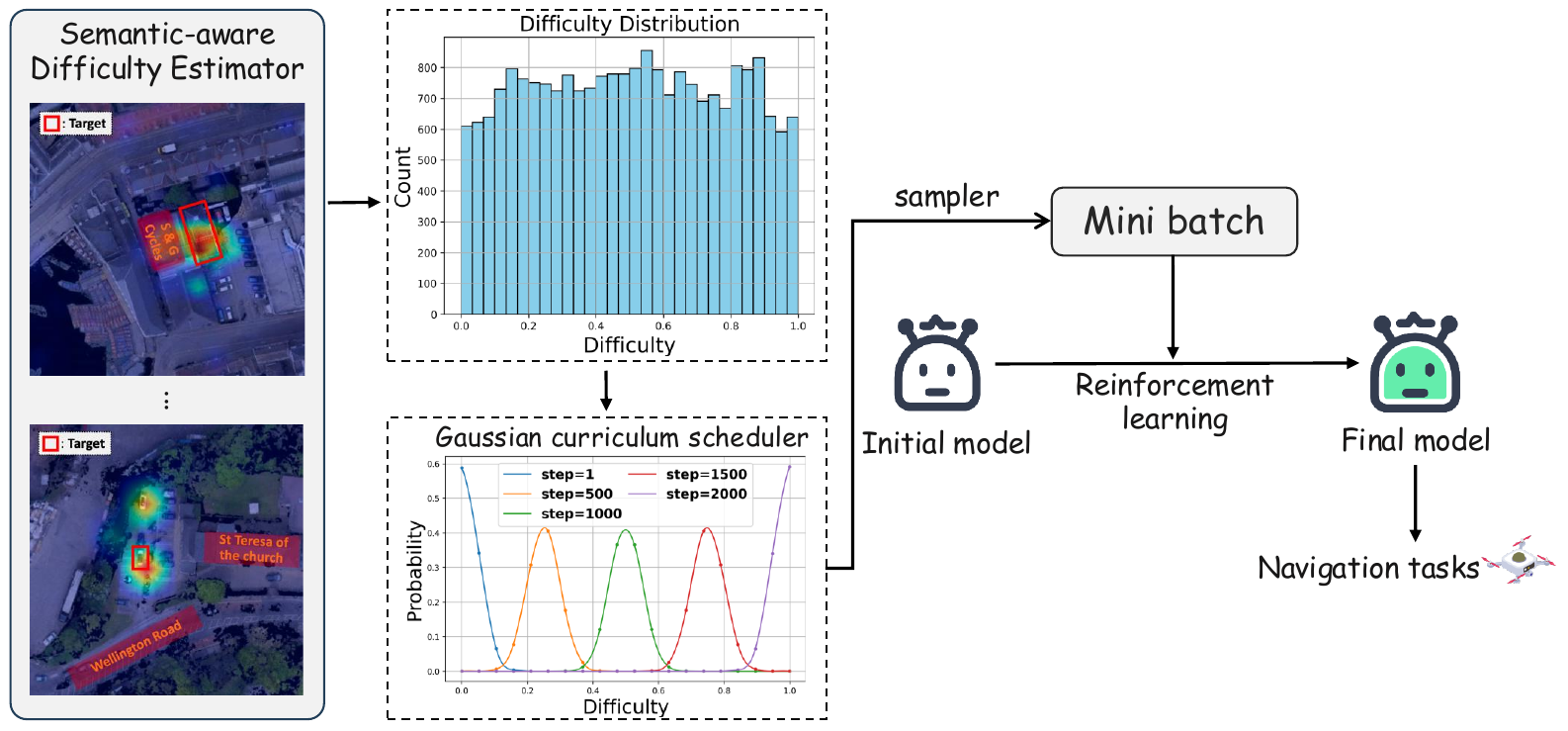} 
\caption{Semantic-Aware Gaussian Curriculum Scheduling (SA-GCS) framework.
A Semantic-Aware Difficulty Estimator (SA-DE) assigns difficulty scores to training samples.
A Gaussian Curriculum Scheduler (GCS) then dynamically adjusts the sampling probabilities based on difficulty scores, enabling the sampler to construct mini-batches that facilitate gradual learning.
The model is optimized via reinforcement learning, enabling a smooth transition from easy to hard tasks, and is ultimately applied to UAV VLN.}
\label{figure:intro}
\end{figure}

To address these issues, Curriculum Learning (CL) has been introduced to RL training to improve stability, efficiency, and performance. 
Inspired by the human learning process, CL starts from easier tasks and gradually progresses to harder ones to establish a more robust learning trajectory. In the context of UAV VLN, this implies that agents should first learn to navigate using clear instructions and simple maps, and then gradually adapt to complex spatial layouts and ambiguous language expressions, thereby enhancing overall performance and generalization.
Based on this insight, we propose a \textbf{Semantic-Aware Gaussian Curriculum Scheduling (SA-GCS)} framework, which systematically incorporates CL into RL for UAV VLN. 
As illustrated in Figure~\ref{figure:intro}, we first design a Semantic-Aware Difficulty Estimator (SA-DE) to assess the difficulty of training samples. Then, we introduce a Gaussian Curriculum Scheduler (GCS) that dynamically adjusts the sampling distribution during training. 
This distribution maintains a Gaussian shape at each stage, with its mean gradually shifting towards harder samples over time. 
The sampled data form mini-batches for RL training, progressively guiding the agent to evolve into a capable navigator that can interpret instructions and complete UAV VLN tasks in challenging environments.

The main contributions are as follows:
\begin{itemize}
  \item We introduce a novel SA-DE that leverages the cross-modal attention maps of VLMs and the Soft-IoU metric to accurately quantify the alignment between the model's attention and the target region, providing a reliable measure of the model’s confidence and precision in target localization.

  \item We present the SA-GCS framework, which seamlessly integrates SA-DE with a GCS to embed curriculum learning into reinforcement learning. 
  This principled design establishes a smooth easy-to-hard training trajectory, substantially enhancing both training efficiency and performance.

  \item Through comprehensive experiments on the CityNav dataset, our approach consistently surpasses baseline methods across nearly all metrics and demonstrates faster convergence, firmly validating its effectiveness and efficiency for UAV VLN.

  \item We further verify the scalability of SA-GCS across models with different parameter sizes, underscoring its potential as a general and reliable training paradigm.
\end{itemize}

%% file: chapters/2_related_work.tex
\section{Related Work}

% % Related Work start

\subsection{Vision-Language Navigation: Supervised and Reinforcement Learning Approaches}
UAV VLN tasks require an agent to navigate in a visually perceived environment by interpreting natural language instructions and integrating multi-view visual inputs. The objective is to perform target localization and flight path planning, making it a representative task that combines cross-modal understanding with autonomous decision-making \cite{wang2024towards}.

Early works in UAV VLN predominantly adopted supervised learning frameworks to optimize performance.
For instance, Seq2Seq models encode image and instruction sequences to produce outputs, marking an initial exploration of UAV VLN \cite{anderson2018vision}.
Building upon this, CMA mechanisms were introduced to enhance perception–language alignment\cite{liu2023aerialvln}. 
Further advancements such as the MGP integrated language understanding and visual perception to construct navigation maps and perform goal prediction, demonstrating improved structural reasoning and planning abilities \cite{lee2024citynav}.

With the recent advances in LLMs and VLMs, researchers have started leveraging general-purpose multimodal models for UAV VLN. 
Benefiting from their strong cross-modal perception capabilities, these models can achieve performance comparable to task-specific systems even with limited training data \cite{saxena2025uav, liu2024navagent}. 
However, their generalization capability in complex and unseen environments remains limited, hindering practical deployment.

To address this, RL has been explored as a post-training strategy to improve generalization. For example, FlightGPT proposed a two-stage training pipeline consisting of SFT followed by RL, and demonstrated its effectiveness on UAV VLN \cite{cai2025flightgpt}. 
Recent studies, including those from the Google team, further validated that RL-based post-training outperforms pure SFT in enhancing model generalizability \cite{chu2025sft, tan2025reason}.

Nevertheless, RL approaches still face challenges such as unstable reward signals, slow convergence, and low sample efficiency.
Designing robust and efficient training mechanisms remains an open and urgent problem in this area.

\subsection{Curriculum Learning in Reinforcement Learning}
CL is a training paradigm that aims to improve learning efficiency and generalization by organizing tasks in a sequence from easy to hard~\cite{bengio2009curriculum}. In the context of RL, CL has been widely adopted to address challenges such as reward sparsity and training instability. By starting with easier tasks or samples, the agent can receive timely feedback and establish basic policies, which can then be gradually extended to more complex problem spaces.

In existing research, many approaches explicitly control task difficulty to guide the training process. For example, Guided Curriculum Learning for Walking Over Complex Terrain~\cite{tidd2020guided} designs a three-stage curriculum with increasing difficulty in terrain and external perturbations, enabling robots to progressively learn from flat ground to challenging obstacles, significantly reducing convergence time and improving success rates. 
Reverse Curriculum Generation~\cite{florensa2017reverse} initiates training from goal states and gradually expands to more distant starting positions, guiding agents to approach the goal from easy to hard, which accelerates learning in tasks like navigation and object grasping.
% Other approaches use heuristic rules to dynamically select training samples. For instance, Success-Induced Task Prioritization (SITP)~\cite{nesterova2022reinforcement} adjusts sampling probabilities based on an agent’s task-specific success rates, favoring samples with intermediate difficulty to balance exploration and exploitation, thereby improving sample efficiency in multi-task control benchmarks.

Moreover, to support smoother transitions between difficulty levels, recent studies introduce strategies based on reward shaping and distribution-based sampling. Adaptive Curriculum Reinforcement Finetuning (AdaRFT)~\cite{shi2025efficient} dynamically adjusts the difficulty of mathematical reasoning problems in LLM finetuning based on real-time reward signals, reducing training time by half and significantly improving accuracy. 
The Easy-to-Hard Gaussian Scheduler (E2H-G)~\cite{parashar2025curriculum} models the sampling probability of tasks as a Gaussian distribution that shifts over training steps, leading to improved performance and sample efficiency for small-scale models in RL-based LLM finetuning.

% Despite the demonstrated effectiveness of CL-RL approaches in mathematical reasoning, robotic control, and even complex locomotion, there is a lack of systematic exploration in the context of UAV VLN—a multimodal task that requires semantic alignment between images and language, spatial reasoning, and action planning. 
% These requirements impose additional challenges on the design of curriculum strategies. 
% To fill this gap, we propose a SA-GCS framework.
% Our method leverages attention heatmaps and Soft-IoU to estimate sample difficulty and adopts a Gaussian Curriculum Scheduler that evolves with training progress.
% This design enables dynamic sample selection and facilitates both training efficiency and model performance in UAV VLN tasks.

Despite these advances, curriculum-based RL has not been systematically explored in UAV VLN, which involves unique challenges such as multimodal semantic alignment, spatial reasoning, and sequential decision-making. 
To bridge this gap, we propose the SA-GCS framework, which leverages cross-modal attention heatmaps of VLMs and Soft-IoU for difficulty estimation, and applies a GCS to dynamically guide training.
This enables efficient sample selection and improves both convergence and generalization in UAV VLN.

%% file: chapters/3_method.tex
\section{Method}

\subsection{Task Formulation}
\label{sec:task_formulation}
This work investigates VLN for UAVs, which aims to guide a UAV to reach a designated target location in a complex environment based on natural language instructions. Specifically, we follow the standard task formulation introduced in the CityNav benchmark~\cite{lee2024citynav}, wherein each navigation instance is represented as a triplet $(I, D, E)$, defined as follows:
\begin{itemize}
    \item $I$ denotes the initial state of the UAV, including its starting position and orientation;
    \item $D$ is a natural language instruction that typically describes the characteristics of the target location and its surrounding landmarks;
    \item $E$ represents the navigation environment, which contains realistic spatial structures and rich semantic elements such as roads, buildings, and vehicles.
\end{itemize}
During navigation, the UAV receives first-person RGB images and depth maps from the environment, along with access to a two-dimensional map aligned with real-world urban geoinformation.
To complete the navigation task, the UAV executes a sequence of discrete actions, including \textbf{move forward}, \textbf{turn left}, \textbf{turn right}, \textbf{ascend}, \textbf{descend}, and \textbf{stop}. When the UAV determines that it is sufficiently close to the target location, it may choose to execute the \textbf{stop} action. A navigation episode is considered successful if the final position of the UAV lies within a predefined threshold (e.g., 20 meters) from the ground-truth target location.

\subsection{System Framework and Reinforcement Learning}
To accomplish the task described in Section~\ref{sec:task_formulation}, we follow the technical pipeline adopted in prior work~\cite{anderson2018vision, liu2023aerialvln, lee2024citynav}, which consists of four sequential modules: \textbf{Perception Acquisition}, \textbf{Goal Inference}, \textbf{Action Planning}, and \textbf{Environment Interaction}.
Furthermore, we introduce reinforcement learning to optimize the performance and generalization of the \textbf{Goal Inference} module.

\subsubsection{System Pipeline}
\begin{figure}[t]
\centering
\includegraphics[width=0.47\textwidth]{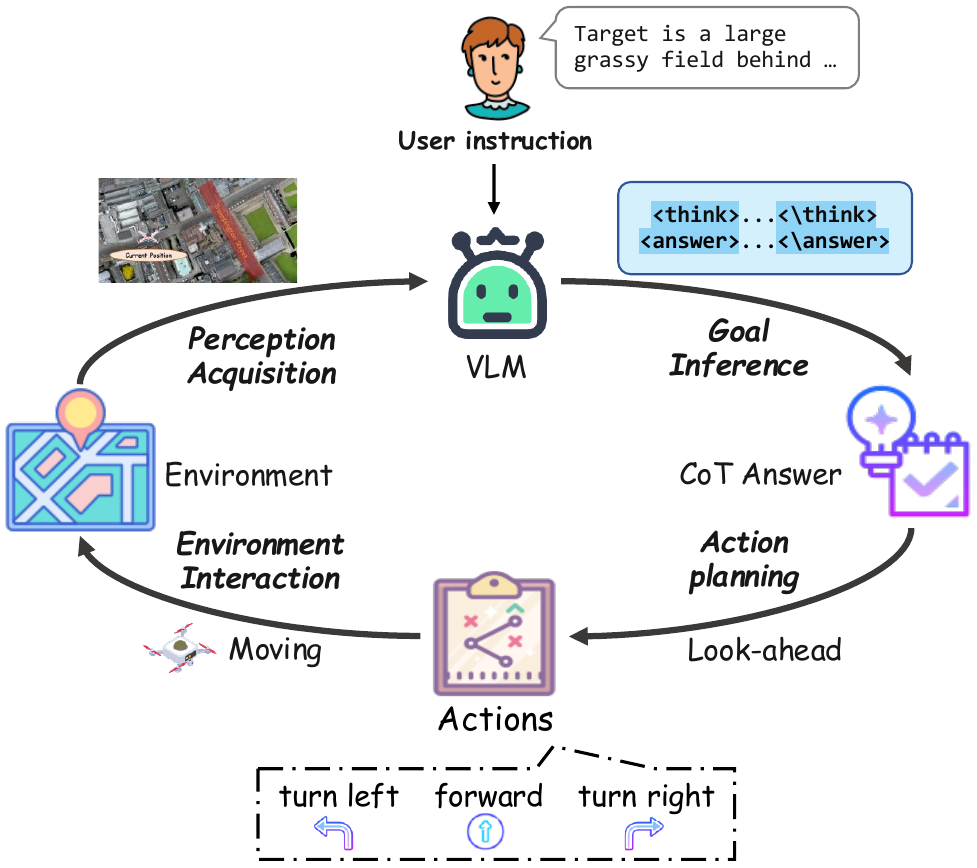} 
\caption{The system pipeline for UAV VLN.}
\label{figure:pipeline}
\end{figure}

As illustrated in Figure~\ref{figure:pipeline}, the system operates in a closed-loop manner, consisting of the following components:
\begin{enumerate}
    \item \textbf{Perception Acquisition} \\
    The system collects two types of real-time input information:
    \begin{itemize}
        \item \textit{Semantic Map:} A structured representation integrating key geographical semantic elements such as roads, buildings, and landmarks. It also encodes the UAV's current spatial position and orientation.
        \item \textit{Natural Language Instruction:} A textual description outlining the characteristics of the target and its surrounding reference objects.
    \end{itemize}
    \item \textbf{Goal Inference} \\
    Based on the input, the model generates a structured reasoning chain that progressively extracts clues related to the target. It then predicts the final coordinates of the target on the map.
    \item \textbf{Action Planning} \\
    Once the target location is determined, the system adopts a look-ahead mechanism~\cite{liu2023aerialvln}, which estimates a short-term reference trajectory from the current position and selects the optimal next action accordingly.
    \item \textbf{Environment Interaction} \\
The UAV executes the planned action, resulting in updates to both the environment and its own state.
\end{enumerate}
This loop is repeated continuously until a \textbf{stop} action is triggered or the maximum number of iterations is reached. This entire pipeline aligns with existing methodologies, ensuring both fair comparison and reproducibility across different approaches.

\subsubsection{Reinforcement Learning Mechanism}
To enhance the performance and generalization capability of the \textbf{Goal Inference} module in dynamic environments, we adopt the Group Relative Policy Optimization (GRPO) algorithm for policy training. 
% The model takes multimodal inputs and generates a structured reasoning process, ultimately outputting a prediction of the target location.
During the training, we adopt a composite reward function consisting of three components, following the reward design proposed in FlightGPT \cite{cai2025flightgpt}, to guide the model's learning from multiple perspectives.
\begin{itemize}
    \item \textbf{Goal Accuracy Reward}: Measures the Euclidean distance between the predicted target coordinates and the ground-truth location, encouraging higher positional accuracy in target prediction.
    \item \textbf{Intermediate Reasoning Reward}: Computes the Intersection-over-Union (IoU) between the predicted landmark region during the \texttt{<think>} phase and the actual landmark area, guiding the model to focus on key semantic cues.
    \item \textbf{Format Reward}: Assesses whether the reasoning output conforms to a predefined structural format, encouraging the generation of well-organized and logically coherent intermediate steps.
\end{itemize}
The final reward is the sum of the three components and is used to update the policy network. 
% This reinforcement mechanism not only improves the model's performance in goal inference but also lays the foundation for the integration of curriculum learning strategies in subsequent training stages.

\subsection{Semantic-Aware Difficulty Estimator}
\label{sec:sa-difficulty-estimator}
In UAV VLN tasks, different samples exhibit significant variations in terms of language ambiguity, visual scene complexity, and the degree of semantic alignment between language instructions and navigation targets. If all training samples are used indiscriminately during training, the model may overfit on easy samples while underfitting on complex ones. 
This imbalance can degrade model performance, weaken generalization ability, and slow down convergence.
To address this issue, we propose a SA-DE that enables automated identification of sample difficulty. Specifically, we leverage cross-modal attention maps produced by a VLM during image-text matching. The alignment between these attention maps and the ground-truth target regions serves as an indicator of semantic understanding. By measuring the degree of correspondence, we can quantitatively assess how difficult a sample is for the model to interpret and locate semantically relevant regions.

\subsubsection{Cross-Modal Attention Heatmap Construction}
\begin{figure}[t]
\centering
\includegraphics[width=0.47\textwidth]{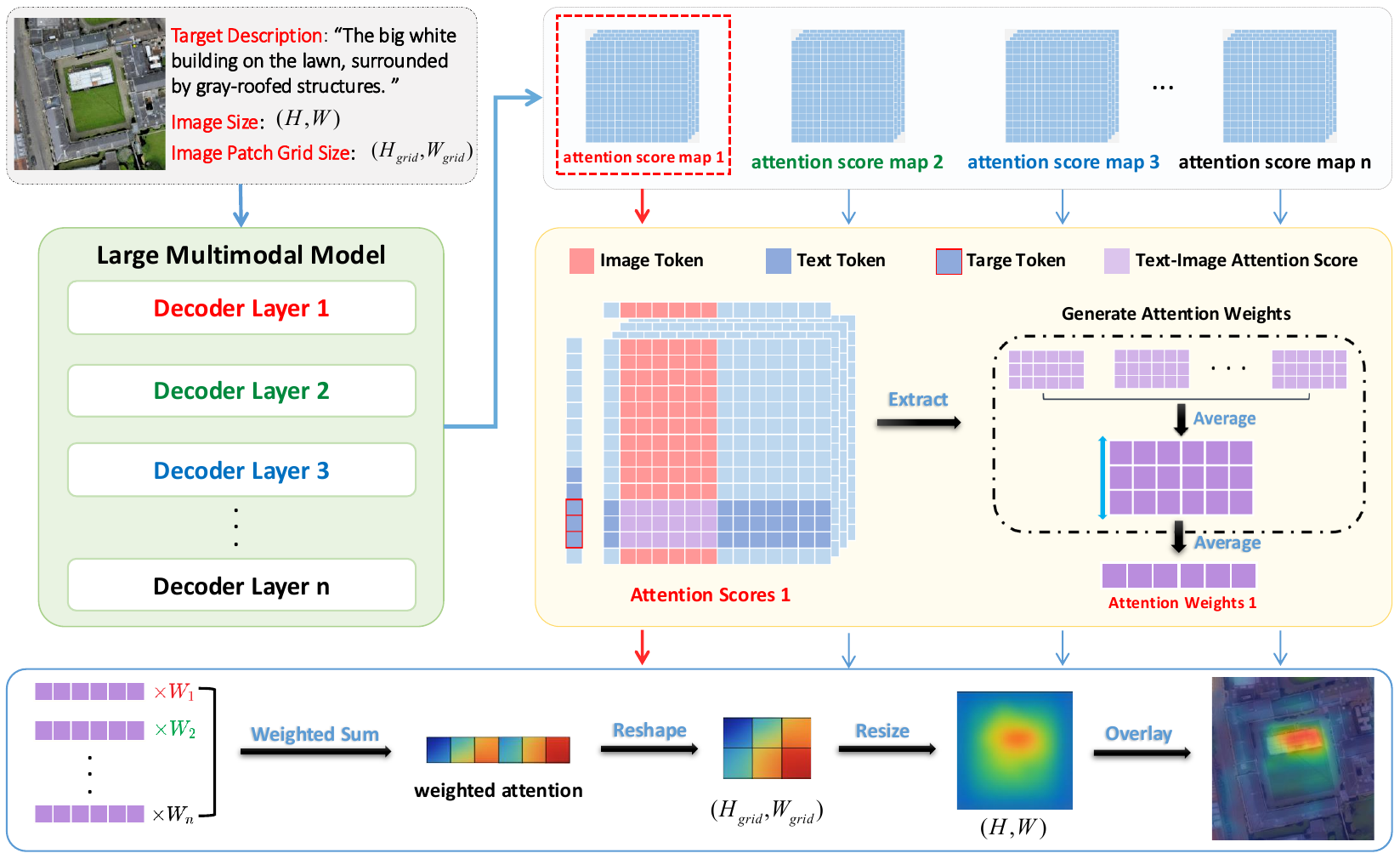}
\caption{Workflow for extracting and visualizing cross-modal attention maps from a Vision-Language Model.}
\label{figure:cross-attention}
\end{figure}

To analyze the model's attention over image regions guided by natural language instructions, we design a cross-attention-based visualization pipeline that generates interpretable attention heatmaps.
These heatmaps offer a clear and interpretable visualization of the model’s attention allocation over image regions, revealing its underlying mechanisms in vision-language matching.
% These heatmaps provide an intuitive representation of the model's attention distribution during vision-language matching.
This approach is inspired by prior works such as BLIP-2~\cite{li2023blip}, GIT~\cite{wang2022git}, and Q-GroundCAM~\cite{rajabi2024q}, which demonstrate that cross-modal attention in Transformer architectures serves as an effective and interpretable signal for assessing semantic alignment.
As illustrated in Figure~\ref{figure:cross-attention}, the generation of attention heatmaps consists of the following steps:
\paragraph{1. Extract Cross-Modal Attention Weights}
Given a vision-language input pair, we feed it into the VLM and extract the cross-attention matrices from each decoder layer. These matrices capture the attention distribution of each text token over the visual patches.
\paragraph{2. Locate Target-Relevant Tokens}
We identify the specific content in the instruction that describes the target (e.g., red building) and determine the corresponding text tokens. We then aggregate their attention weights over the image tokens.
\paragraph{3. Fuse Attention Across Layers}
% To integrate multi-layer attention signals, we apply a weighted averaging strategy, where higher layers are assigned greater weights. We further average across all attention heads to obtain a final one-dimensional attention vector.
We apply a weighted averaging strategy over multi-layer attention signals, assigning greater weights to higher layers and averaging across heads to produce the final one-dimensional attention vector.
\paragraph{4. Generate and Visualize Heatmap}
The resulting attention vector is projected back onto the image patch grid, upsampled to the original image resolution via interpolation, and overlaid on the input image to visualize the model’s attention regions.

\subsubsection{Difficulty Score Computation}
To quantitatively assess the alignment between the Cross-Modal Attention Map and the actual target region, we adopt the Soft-IoU metric~\cite{jiang2018acquisition}. 
Let $M \in \{0, 1\}^{H \times W}$ denote the binary mask representing the ground-truth target region, and let $H \in [0, 1]^{H \times W}$ denote the attention heatmap generated by the model.
The Soft-IoU is defined as:
\[
\text{IoU}_{\text{soft}} = \frac{\sum (H \cdot M)}{\sum H + \sum M - \sum (H \cdot M)}
\]
Based on this, we define the Semantic-Aware Difficulty score as:
\[
\text{Difficulty} = 1 - \text{IoU}_{\text{soft}}, \quad \text{Difficulty} \in [0, 1].
\]
This metric quantitatively reflects both the model’s localization confidence and its alignment accuracy with the ground-truth target region.
A score closer to 1 indicates low confidence and misaligned attention, suggesting the sample is semantically harder for the model. 
Conversely, a score near 0 implies accurate focus and high confidence, indicating an easier sample. 
Thus, the difficulty can serve as a vital reference in curriculum learning.
The distribution of sample difficulty in the training set is provided in Appendix~\ref{appendix:difficulty_distribution} for reference.

\begin{figure}[t]
\centering
\includegraphics[width=0.47\textwidth]{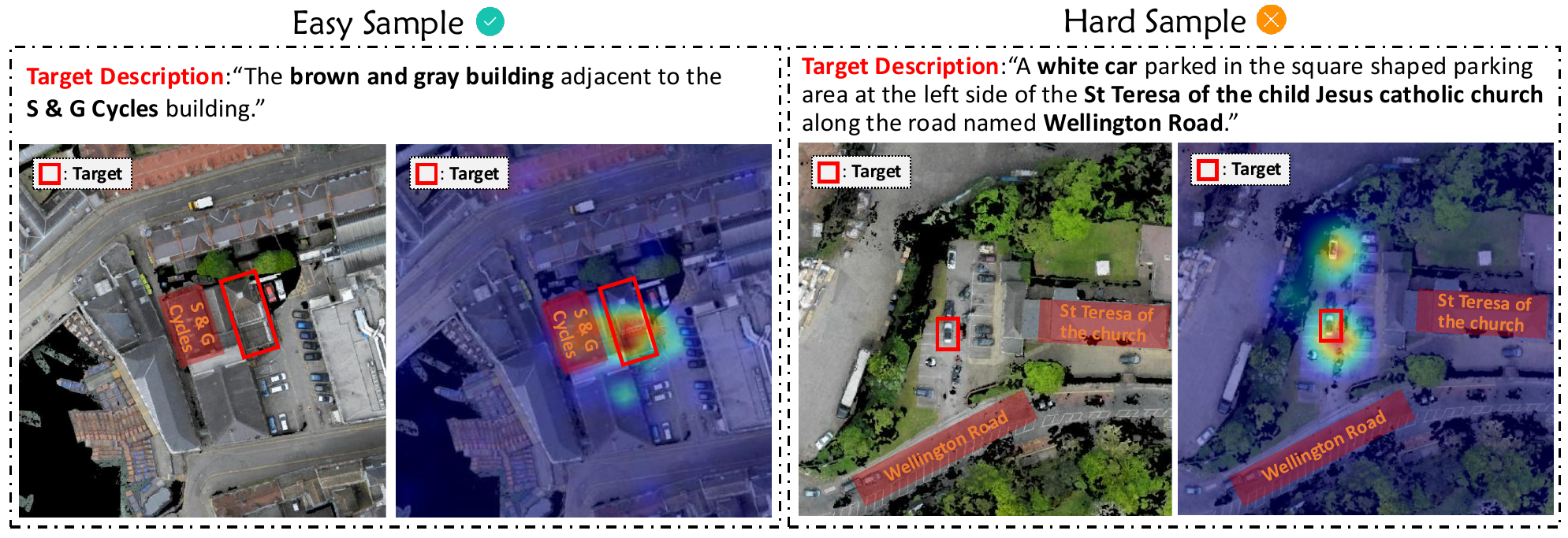} 
\caption{Representative examples of easy and hard samples in difficulty estimation.}
\label{figure:easy_hard_sample}
\end{figure}

Figure~\ref{figure:easy_hard_sample} illustrates two representative cases. 
The \textbf{easy sample} (left) corresponds to the instruction: \emph{``the brown and gray building adjacent to the S \& G Cycles building''}. 
In this case, the target building is located immediately next to a clearly labeled landmark (“S \& G Cycles”), offering strong visual and spatial cues. 
The model's attention is highly concentrated on the correct region, resulting in a heatmap that overlaps significantly with the ground-truth mask. 
Consequently, the Soft-IoU score is high and the computed difficulty is low, indicating that the model successfully focuses on and identifies the target.
In contrast, the \textbf{hard sample} (right) is associated with a more complex instruction: \emph{``a white car parked in the square-shaped parking area at the left side of the St Teresa of the Child Jesus catholic church along the road named Wellington Road''}. This description involves multiple reference points and relative spatial relationships, introducing semantic ambiguity. 
Furthermore, the image contains several white vehicles that could serve as potential candidates. As a result, the attention heatmap is dispersed and overlaps poorly with the ground-truth region, leading to a lower Soft-IoU score and a higher difficulty value.
% This reflects the increased challenge that this sample poses for the model’s perception and localization capabilities.
This underscores the significant challenge that this sample imposes on the model’s perception and localization capabilities.

\subsection{Semantic-Aware Gaussian Curriculum Scheduling Framework}
Although RL has demonstrated promising generalization capabilities in UAV VLN, it often suffers from slow convergence, low sample efficiency, and training instability. To address these issues, we propose a novel training framework named SA-GCS. This framework systematically integrates CL into the RL process by dynamically adjusting the training focus from easy to hard samples.

To achieve this, we design a GCS that controls the sampling probability distribution over training samples at each step. 
% Specifically, the sampling distribution is modeled as a Gaussian {\color{red} process} whose mean gradually shifts toward harder samples as training progresses. 
Specifically, the sampling distribution is modeled as a Gaussian distribution whose mean gradually shifts toward harder samples as training progresses.
This design enables a smooth and progressive learning trajectory from simple to complex tasks, thereby improving both training efficiency and final performance.

\textbf{Gaussian Curriculum Scheduler.}
We first assign a Semantic-Aware Difficulty score $d_i \in [0, 1]$ to each training sample using the SA-DE introduced in Section~\ref{sec:sa-difficulty-estimator}.
At training step $t$, the scheduler generates a Gaussian distribution based on the current training progress, and computes the sampling probability $P_i(t)$ for each sample as:
\begin{equation}
P_i(t) = \frac{1}{Z} \exp\left(-\frac{(d_i - \mu(t))^2}{2\sigma^2}\right).
\end{equation}
Here, $\mu(t) \in [0, 1]$ is the mean of the Gaussian distribution, which gradually shifts from lower to higher difficulty levels as training progresses. The parameter $\sigma$ controls the spread (standard deviation) of the distribution, and $Z$ is a normalization constant ensuring that all sampling probabilities sum to 1.
$\mu(t)$ is updated dynamically via a linear scheduling strategy:
\begin{equation}
\mu(t) = \mu_0 + \frac{t}{T}(1 - \mu_0),
\end{equation}
where $\mu_0$ denotes the initial mean value, and $T$ is the total number of training steps. 
% This scheduling mechanism allows the model to focus on easier samples in the early stages and gradually shift attention to harder ones, achieving a smooth and progressive learning trajectory.

The overall SA-GCS process is summarized in Algorithm~\ref{alg:sa-gcs}.
\begin{algorithm}[H]
\caption{Semantic-Aware Gaussian Curriculum Scheduling}
\label{alg:sa-gcs}
\KwIn{
    Difficulty scores $\{d_i\}_{i=1}^N$ for training samples;\\
    Total training steps $T$;\\
    Initial mean $\mu_0$ and standard deviation $\sigma$.
}
\KwOut{Final trained reinforcement learning model.}

Initialize the reinforcement learning model\;

\For{$t = 1$ \KwTo $T$}{
    Update the current mean $\mu(t)$\;
    
    \ForEach{sample $i$}{
        Compute sampling probability $P_i(t)$\;
    }

    Sample a mini-batch from the dataset based on $P_i(t)$\;
    
    Perform policy optimization using the mini-batch\;
    
    Update model parameters\;
}

\Return final trained model\;
\end{algorithm}

Compared to static sampling methods, the proposed SA-GCS provides several benefits:
\begin{itemize}
  \item \textbf{Progressive Training:} The learning process transitions smoothly from easy to hard samples, effectively mitigating the sparse reward problem in early stages;
  \item \textbf{Sample Diversity:} The long-tail characteristic of the Gaussian distribution maintains sample diversity at each stage, encouraging broader exploration and reducing overfitting;
  \item \textbf{Resistance to Catastrophic Forgetting:} The scheduler maintains a non-zero probability of sampling easier examples throughout training, enabling the model to reinforce and retain previously learned behaviors over time.
  % \item \textbf{High Adaptability:} The training dynamics can be flexibly tuned through only two hyperparameters ($\mu_0$ and $\sigma$), offering strong generalizability across different models and tasks.
\end{itemize}

% Compared to static sampling methods, the proposed SA-GCS provides several benefits: 
% \textbf{Progressive Training} — the learning process transitions smoothly from easy to hard samples, effectively mitigating the sparse reward problem encountered in early stages; 
% \textbf{Sample Diversity} — the long-tail characteristic of the Gaussian distribution maintains sample diversity at each stage, encouraging broader exploration and reducing overfitting; 
% \textbf{Resistance to Catastrophic Forgetting} — the scheduler maintains a non-zero probability of sampling easier examples throughout training, enabling the model to reinforce and retain previously learned behaviors over time.

%% file: chapters/4_experiments.tex
\section{Experiments}

\subsection{Datasets and Metrics}
We conduct our experiments on the CityNav dataset~\cite{lee2024citynav}, a large-scale simulated benchmark designed for UAV VLN across multiple real-world cities. 
CityNav reconstructs urban layouts, architectural structures, and semantic landmarks within a simulated environment, providing a realistic testbed for navigation systems.
The dataset contains over 90,000 human-written natural language instructions, covering a diverse range of linguistic styles and navigation objectives. 
% Each instruction is paired with a corresponding ground-truth navigation trajectory, making it well-suited for both training and evaluation of UAV VLN models.
Each instruction is paired with a corresponding ground-truth navigation trajectory, making it well-suited for both training and evaluation in UAV VLN.

For performance evaluation, we follow the official metrics defined in the benchmark, which include four metrics:
\begin{itemize}
  \item \textbf{Navigation Error (NE):} The Euclidean distance (in meters) between the UAV's final position and the target location. A lower NE indicates higher accuracy.
  \item \textbf{Success Rate (SR):} The percentage of episodes where the UAV ends within 20 meters of the target location. This reflects the overall task completion rate.
  \item \textbf{Oracle Success Rate (OSR):} The proportion of episodes where any point along the UAV's trajectory comes within 20 meters of the target. This captures whether the target was ever approached during navigation.
  \item \textbf{Success weighted by Path Length (SPL):} Among successful episodes, SPL evaluates the efficiency of the navigation path. It penalizes unnecessary detours and rewards trajectories close to the shortest path.
\end{itemize}
% Together, these metrics offer a comprehensive evaluation of model performance in terms of accuracy, robustness, and efficiency in navigation.

\subsection{Baselines}
We select a set of representative baseline methods for comprehensive comparison, including:
(1) rule-based policy strategies (Random), 
(2) supervised learning methods (Seq2Seq, CMA, MGP), and 
(3) recently proposed multimodal large language models (Qwen, LLaMA, GPT-4o). 
Descriptions and implementation details for these models are provided in Appendix~\ref{appendix:baselines}.

In addition to the baselines, we compare three sampling strategies during RL: Random Sampling, Naive CL, and SA-GCS. 
Random Sampling uniformly selects training samples without considering their difficulty. 
Naive CL ranks all samples by their difficulty estimated by the SA-DE module and enforces an easy-to-hard training schedule.

% Naive CL ranks all samples by their difficulty estimated by the SA-DE module and enforces an easy-to-hard training schedule, but it lacks an adaptive probabilistic scheduling mechanism.

% In addition to the baselines, we also compare three sampling strategies during RL: 
% Random Sampling, Naive CL, and our proposed SA-GCS. 
% Random Sampling selects training samples uniformly without considering difficulty. 
% Naive CL sorts all samples by Semantic-Aware Difficulty and trains the model in an easy-to-hard order, 
% but does not include any dynamic adjustment or probabilistic sampling like SA-GCS.

\subsection{Implement Details}
We conduct our experiments based on the Qwen2.5-VL-7B model, applying the GRPO algorithm for policy learning. In addition to this standard training pipeline, we design a comparative setting where the model is first warm-started using SFT before undergoing reinforcement learning. This SFT + RL training paradigm has been shown to be effective in prior studies~\cite{cai2025flightgpt}.
A customized prompt template is designed to guide the model in producing structured reasoning and accurate target predictions, with its complete format detailed in Appendix~\ref{appendix:prompt}.
The hyperparameters for SFT and RL are listed in Table~\ref{tab:sft-rl-hyperparams}, while those for CL are summarized in Table~\ref{tab:cl-hyperparams}.

\begin{table}[t]
\small
\centering
\caption{Hyperparameters for SFT and RL}
\label{tab:sft-rl-hyperparams}
\begin{tabularx}{\columnwidth}{lXXX}
\toprule
Stage & Batch Size & Learning Rate & \#Epochs \\
\midrule
SFT & 16 & 2e-5 & 2 \\
RL  & 1 & 1e-5 & 1 \\
\bottomrule
\end{tabularx}
\end{table}

\begin{table}[t]
\small
\centering
\caption{Hyperparameters for CL}
\label{tab:cl-hyperparams}
\begin{tabularx}{\columnwidth}{lXXX}
\toprule
\textbf{$\mu_0$} & \textbf{$\sigma$} & \#Steps & Samples/Step \\
\midrule
0 & 0.3 & 2000 & 2 \\
\bottomrule
\end{tabularx}
\end{table}

%% file: chapters/5_results_and_analysis.tex
\section{Results and Analysis}

\subsection{Overall Performance}

Table~\ref{tab:all_result} presents a comparative analysis of our proposed method against several representative baselines on the CityNav dataset, under three data split settings and four evaluation metrics. The key findings are summarized as follows.

\textbf{CL consistently improves model performance while maintaining strong generalization.}
Both the Naive CL and the SA-GCS outperform the Random sampling baseline across almost all evaluation metrics and data splits. 
These results confirm the effectiveness of CL in the UAV VLN,  and validate the rationality of the proposed SA-DE.

\textbf{The SA-GCS achieves the best overall performance.}
Among all evaluated methods, the SA-GCS achieves the strongest results across almost all evaluation metrics.
Specifically, on the \textit{Test-Unseen} split, it achieves a SR of 24.55\% and a SPL score of 22.86\%, outperforming both the Naive CL (SR 23.03\%, SPL 21.36\%) and the Random sampling strategy (SR 21.06\%, SPL 19.21\%). 
These results highlight the superiority of the SA-GCS in guiding sample selection and controlling training progression.

\textbf{The SA-GCS is robust across different model initialization settings.}
Whether applied to the original model without SFT or used after an SFT-based cold start, the SA-GCS consistently enhances performance. This demonstrates the proposed training framework's strong generalizability and adaptability to different model states.

% \begin{itemize}
% \item \textbf{Curriculum learning consistently improves model performance while maintaining strong generalization.} \\
% Both the Naive strategy and the Gaussian strategy outperform the random sampling baseline across all evaluation metrics and data splits. These results confirm the effectiveness of curriculum learning in the UAV VLN task,  and validate the rationality of the proposed semantic-aware difficulty estimation.
% \item \textbf{The Gaussian curriculum strategy achieves the best overall performance.} \\
% Among all evaluated methods, the Gaussian curriculum strategy achieves the strongest results across almost all evaluation metrics. Specifically, on the \textit{Test-Unseen} split, it achieves a success rate (SR) of 24.55\% and a SPL score of 22.86\%, outperforming both the Naive curriculum (SR 23.03\%, SPL 21.36\%) and the Random sampling strategy (SR 21.06\%, SPL 19.21\%). These results highlight the superiority of the Gaussian scheduler in guiding sample selection and controlling training progression.
%  \item \textbf{The curriculum strategy is robust across different model initialization settings.} \\
% Whether applied to the original model without supervised fine-tuning (SFT) or used after an SFT-based cold start, the curriculum learning strategy consistently enhances performance. This demonstrates the proposed training framework's strong generalizability and adaptability to different model states.
% \end{itemize}

\begin{figure}[t]
\centering
\includegraphics[width=0.47\textwidth]{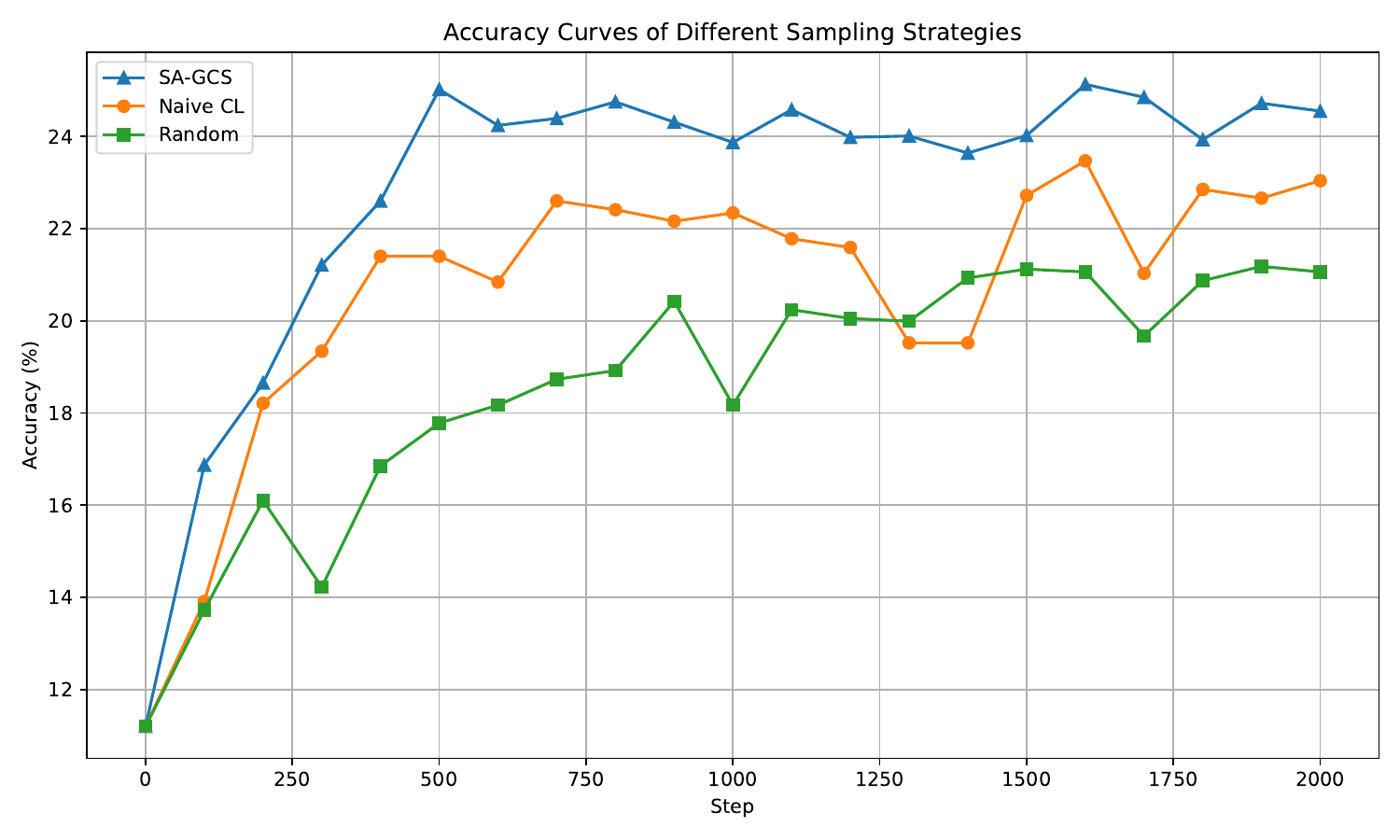} 
\caption{Success Rate progression on the Test-Unseen split using different sampling strategies.}
\label{figure:step-sr}
\end{figure}

\begin{table*}[ht]
\centering
\tiny
\setlength{\tabcolsep}{2pt}
\caption{Comparison of Model Performance Across Different Data Splits.}
\label{tab:all_result}
\begin{tabular}{>{\centering\arraybackslash}p{3.7cm} | >{\centering\arraybackslash}p{0.77cm} >{\centering\arraybackslash}p{0.77cm} >{\centering\arraybackslash}p{0.77cm} >{\centering\arraybackslash}p{0.77cm} | >{\centering\arraybackslash}p{0.77cm} >{\centering\arraybackslash}p{0.77cm} >{\centering\arraybackslash}p{0.77cm} >{\centering\arraybackslash}p{0.77cm} | >{\centering\arraybackslash}p{0.77cm} >{\centering\arraybackslash}p{0.77cm} >{\centering\arraybackslash}p{0.77cm} >{\centering\arraybackslash}p{0.77cm}}
\toprule
\multirow{2}{*}{\textbf{Method}} & \multicolumn{4}{c|}{\textbf{Validation Seen}} & \multicolumn{4}{c|}{\textbf{Validation Unseen}} & \multicolumn{4}{c}{\textbf{Test Unseen}} \\
& NE$\downarrow$ & SR$\uparrow$ & OSR$\uparrow$ & SPL$\uparrow$ & NE$\downarrow$ & SR$\uparrow$ & OSR$\uparrow$ & SPL$\uparrow$ & NE$\downarrow$ & SR$\uparrow$ & OSR$\uparrow$ & SPL$\uparrow$ \\
\midrule
Random & 222.30 & 0.00 & 1.15 & 0.00 & 223.00 & 0.00 & 0.90 & 0.00 & 208.80 & 0.00 & 1.44 & 0.00 \\
Seq2Seq & 148.40 & 4.52 & 10.61 & 4.47 & 201.40 & 1.04 & 8.03 & 1.02 & 174.50 & 1.73 & 8.57 & 1.69 \\
CMA & 151.70 & 3.74 & 10.77 & 3.70 & 205.20 & 1.08 & 7.89 & 1.06 & 179.10 & 1.61 & 10.07 & 1.57 \\
MGP & 59.70 & 8.69 & \textbf{35.51} & 8.28 & 75.10 & 5.84 & 22.19 & 5.56 & 93.80 & 6.38 & 26.04 & 6.08 \\
Qwen2.5-VL-32B & 84.70 & 12.65 & 24.14 & 11.30 & 91.90 & 10.12 & 20.52 & 9.00 & 83.28 & 11.98 & 23.48 & 10.76 \\
LLaMA-3.2-11B-Vision & 198.90 & 1.16 & 5.16 & 1.06 & 215.10 & 0.50 & 4.35 & 0.46 & 191.10 & 1.26 & 4.59 & 1.15 \\
GPT-4o & 155.80 & 2.42 & 9.62 & 2.17 & 170.40 & 2.17 & 7.77 & 1.98 & 144.40 & 3.90 & 11.79 & 3.42 \\
FlightGPT & 66.10 & 17.57 & 30.26 & 15.78 & 68.10 & 14.69 & 29.33 & 13.24 & 76.20 & 21.20 & 35.38 & 19.24 \\
Qwen2.5-VL-7B & 116.10 & 4.72 & 12.89 & 4.15 
              & 123.20 & 5.52 & 13.98 & 4.92 
              & 124.60 & 4.59 & 12.75 & 3.99 \\
Qwen2.5-VL-7B + RL (Random) & 74.90 & 13.27 & 27.13 & 12.59 
                          & 71.40 & 12.87 & 25.82 & 12.27 
                          & 76.50 & 19.80 & 32.26 & 18.91 \\
Qwen2.5-VL-7B + RL (Naive CL) & 66.99 & 15.53 & 27.90 & 14.16 
                            & 68.73 & 15.78 & 30.49 & 14.22 
                            & 76.09 & 21.05 & 33.71 & 19.23 \\
Qwen2.5-VL-7B + RL (SA-GCS) & 62.40 & 18.25 & 31.59 & 16.94 
                               & 63.86 & 16.56 & 29.30 & 15.25 
                               & 75.18 & 22.20 & 33.93 & 20.61 \\
Qwen2.5-VL-7B-SFT & 97.60 & 10.29 & 18.45 & 9.46 
                  & 101.70 & 10.51 & 18.54 & 9.70 
                  & 117.40 & 11.20 & 21.24 & 10.78 \\
Qwen2.5-VL-7B-SFT + RL (Random) & 67.75 & 15.54 & 27.22 & 13.98 
                              & 67.58 & 15.49 & 29.01 & 13.90 
                              & 75.32 & 21.06 & 35.09 & 19.21 \\
Qwen2.5-VL-7B-SFT + RL (Naive CL) & 71.19 & 16.93 & 28.26 & 15.81 
                                & 72.47 & 15.86 & 28.02 & 14.76 
                                & 75.61 & 23.03 & 34.59 & 21.36 \\
Qwen2.5-VL-7B-SFT + RL (SA-GCS) & \textbf{59.68} & \textbf{18.69} & 31.98 & \textbf{17.26} & \textbf{63.76} & \textbf{16.88} & \textbf{31.48} & \textbf{15.89} 
                                   & \textbf{68.42} & \textbf{24.55} & \textbf{37.36} & \textbf{22.86} \\
\bottomrule
\end{tabular}
\end{table*}

\subsection{Effectiveness and Efficiency Analysis of SA-GCS}
To verify the effectiveness of the proposed SA-GCS, we further investigate how different sampling strategies influence model performance throughout the training process.
Figure~\ref{figure:step-sr} shows how the SR evolves over training steps on the Test-Unseen split, comparing three strategies initialized from an SFT model: Random Sampling, Naive CL, and SA-GCS.
The results reveal the following key observations.

\textbf{CL accelerates convergence and confirms the effectiveness of the SA-DE.}
Both Naive CL and SA-GCS significantly accelerate the model's convergence compared to Random sampling. 
This finding confirms that incorporating CL effectively enhances training efficiency. Moreover, it validates the rationality of our proposed SA-DE, which successfully distinguishes sample difficulty levels and supports the construction of an easy-to-hard training trajectory.

\textbf{SA-GCS outperforms Naive CL in both convergence speed and final performance.}
The SA-GCS achieves convergence within approximately 500 training steps, with the final SR reaching 24.55\%. 
In contrast, Naive CL requires around 700 steps to converge and attains a slightly lower final SR of 23.03\%. 
% This indicates that the SA-GCS is more efficient in selecting appropriate samples, thereby better facilitating policy optimization.
This indicates that SA-GCS is more efficient in selecting appropriate samples, reducing training cost by about 20 A100 GPU hours and highlighting its practical efficiency.

\textbf{SA-GCS exhibits more stable performance in later training stages, showing stronger resistance to forgetting.}
In the later phase of training (steps 1300–1400), the SR curve of the Naive CL exhibits noticeable fluctuations, suggesting a risk of forgetting earlier learned policies. 
Conversely, the SA-GCS maintains relatively stable performance, indicating superior robustness and generalization under prolonged training.

\subsection{Scalability across Model Sizes}
To evaluate the scalability of the proposed SA-GCS framework across different model sizes, we replace the base model with a smaller version, Qwen2.5-VL-3B, and repeat the same experimental setup.
Both the quantitative results and the evolution of Success Rate during training on the Test-Unseen split are presented in Appendix~\ref{appendix:3b_result}.
Based on the quantitative results and the Success Rate curves over training steps, we draw the following conclusions.

\textbf{CL remains effective for smaller models}: On the 3B model, both Naive CL and SA-GCS consistently outperform the Random sampling baseline, demonstrating the general applicability of the CL mechanism.

\textbf{SA-GCS yields superior performance and generalization}: Across all three data splits, the SA-GCS achieves the best performance across all metrics and simultaneously demonstrates strong generalization ability.

\textbf{More efficient and stable training process}: As shown in the SR training curves, the SA-GCS leads to faster convergence in the early stages and maintains greater stability in later stages. 
This trend is consistent with observations on the 7B model, further validating the transferability and robustness of the proposed method across model scales.

%% file: chapters/6_conclusion.tex
\section{Conclusion}

This paper proposes a SA-GCS framework for UAV VLN. 
The framework integrates a SA-DE to accurately quantify sample difficulty and employs a GCS to dynamically adjust the sampling distribution during training, enabling a progressive learning path from easy to hard samples.
Experimental results on the CityNav dataset demonstrate that SA-GCS achieves state-of-the-art performance across almost all metrics, with faster convergence and stronger generalization capability.
Moreover, the method exhibits stable performance across different model scales, indicating good scalability.
In summary, SA-GCS provides an efficient and robust training paradigm for UAV VLN tasks and serves as a valuable reference for future research in this field.

%% file: chapters/99_appendix.tex
\section{Difficulty Distribution in the Training Set}
\label{appendix:difficulty_distribution}

\begin{figure}[ht]
\centering
\includegraphics[width=0.47\textwidth]{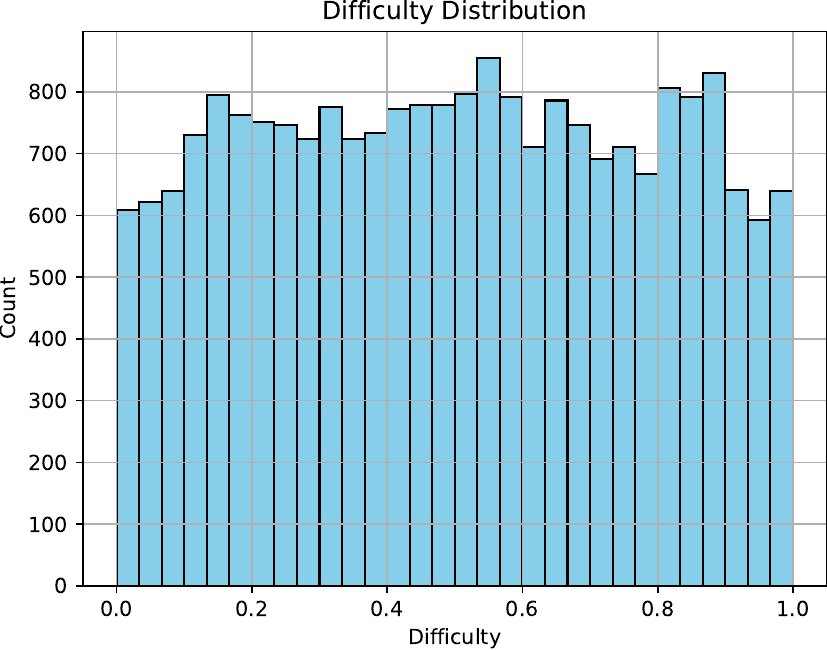} 
\caption{Distribution of difficulty in the training set.}
\label{figure:difficulty_distribution}
\end{figure}

\section{Baseline Method Details}
\label{appendix:baselines}
\begin{itemize}
    \item \textbf{Random} \\
    This baseline randomly selects an action at each time step, without relying on any perceptual input or language instruction. It serves as a reference for strategy-free control performance.
    \item \textbf{Sequence-to-Sequence (Seq2Seq)}~\cite{anderson2018vision}\\
    This method encodes perceptual inputs and language instructions using a recurrent sequence model to produce navigation outputs. 
    It represents an early end-to-end supervised learning approach for navigation tasks.
    \item \textbf{Cross-Modal Attention (CMA)}~\cite{liu2023aerialvln}\\
    Built upon the Seq2Seq framework, CMA introduces a cross-modal attention mechanism to enhance the alignment and fusion of visual and linguistic features, improving the model’s ability to handle multimodal information.
    \item \textbf{Map-based Goal Predictor (MGP)}~\cite{lee2024citynav}\\
    MGP combines a language parser and an object detection module to extract semantic information about goals and landmarks. It then constructs a semantic navigation map to predict the target location.
    \item \textbf{Qwen2.5-VL-7B}~\cite{bai2025qwen25vltechnicalreport}\\
    A multimodal large language model based on the Qwen architecture with 7 billion parameters. It supports joint understanding and reasoning over visual and linguistic inputs.
    \item \textbf{Qwen2.5-VL-32B}~\cite{bai2025qwen25vltechnicalreport}\\
    A larger-scale variant in the Qwen2.5 series, equipped with 32 billion parameters and enhanced capability for complex multimodal reasoning.
    \item \textbf{LLaMA-3.2-11B-Vision}~\cite{grattafiori2024llama}\\
    A vision-language model based on the LLaMA architecture with 11 billion parameters. It effectively handles visual and textual inputs across diverse multimodal tasks.
    \item \textbf{GPT-4o}~\cite{openai2024gpt4technicalreport}\\
    A state-of-the-art multimodal large language model that accepts image and text inputs, and demonstrates strong performance in cross-modal understanding and instruction following.
\end{itemize}

\section{Prompts Template}
\label{appendix:prompt}

\begin{myexample}{UAV VLN task}{}

    \textbf{System Message}: \\
    You are an intelligent autonomous aerial vehicle (UAV) capable of real-world navigation and visual target localization.\\
    
    \textbf{Mission Objective}: \\
    Your mission is to locate a specific target described in natural language instructions. \\
    
    \textbf{Details of the Targe}: \\
    \verb|{target description}| \\
    
    \textbf{Environmental Perception}: \\
    - The UAV's current position is indicated by the starting point of an arrow in the image, with its heading angle represented by the arrow's direction. \\
    - The yellow box outlines the UAV's current camera field of view on the map, centered at pixel coordinates: \verb|cur_pose = {UAV current position}|. \\
    - Landmark regions are highlighted with red masks. \\
    
    \textbf{Operational Guidance}: \\
    - The target is usually located near a red-masked landmark. \\
    - Use both the target description and the visual input to identify the most relevant red-masked landmark region. \\
    - Infer the relative position of the target with respect to that landmark. \\

    \textbf{Output Format Specification}: \\
    - Present your reasoning process within \texttt{<think>} and \texttt{</think>} tags.\\ 
    - Provide your final answer within \texttt{<answer>} and \texttt{</answer>} tags in the following format: \verb|{"target_location": [x, y]}| \\
    Your reasoning may include: \\
          $\quad$- A semantic interpretation of the target description. \\
          $\quad$- Identification of the correct landmark region. \\
          $\quad$- The bounding box of that region in the following format: \\
          \verb|{"landmark_bbox": [x1, y1, x2, y2]}|
\end{myexample}

\section{Results on Qwen2.5-VL-3B Model}
\label{appendix:3b_result}

\begin{table*}[ht]
\centering
\tiny
\setlength{\tabcolsep}{2pt}
\caption{Comparison of Model Performance Across Different Data Splits (3B).}
\label{tab:3b_result}
\begin{tabular}{>{\centering\arraybackslash}p{3.7cm} | >{\centering\arraybackslash}p{0.77cm} >{\centering\arraybackslash}p{0.77cm} >{\centering\arraybackslash}p{0.77cm} >{\centering\arraybackslash}p{0.77cm} | >{\centering\arraybackslash}p{0.77cm} >{\centering\arraybackslash}p{0.77cm} >{\centering\arraybackslash}p{0.77cm} >{\centering\arraybackslash}p{0.77cm} | >{\centering\arraybackslash}p{0.77cm} >{\centering\arraybackslash}p{0.77cm} >{\centering\arraybackslash}p{0.77cm} >{\centering\arraybackslash}p{0.77cm}}
\toprule
\multirow{2}{*}{\textbf{Method}} & \multicolumn{4}{c|}{\textbf{Validation Seen}} & \multicolumn{4}{c|}{\textbf{Validation Unseen}} & \multicolumn{4}{c}{\textbf{Test Unseen}} \\
& NE$\downarrow$ & SR$\uparrow$ & OSR$\uparrow$ & SPL$\uparrow$ & NE$\downarrow$ & SR$\uparrow$ & OSR$\uparrow$ & SPL$\uparrow$ & NE$\downarrow$ & SR$\uparrow$ & OSR$\uparrow$ & SPL$\uparrow$ \\
\midrule
Qwen2.5-VL-3B & 171.16 & 1.48 & 3.36 & 1.44 & 181.64 & 1.17 &
3.36 & 1.10 & 165.56 & 1.45 & 2.90 & 1.41 \\
Qwen2.5-VL-3B + RL (Random) & 79.84 & 11.41 & 23.85 & 10.25 & 87.30 &
5.31 & 17.06 & 4.75 & 93.78 & 10.94 & 23.31 & 9.95 \\
Qwen2.5-VL-3B + RL (Naive CL) & 77.15 & 12.05 & 23.26 & 10.89 & 90.07 &
7.21 & 20.13 & 6.50 & 89.43 & 11.71 & 24.27 & 10.67 \\
Qwen2.5-VL-3B + RL (SA-GCS) & \textbf{73.93} & \textbf{12.10} & \textbf{24.49} & \textbf{11.55} & \textbf{83.04} & \textbf{7.64} & \textbf{21.48} & \textbf{6.88} & \textbf{85.58} & \textbf{12.92} & \textbf{25.56} & \textbf{11.68} \\
\bottomrule
\end{tabular}
\end{table*}

\begin{figure}[ht]
\centering
\includegraphics[width=0.47\textwidth]{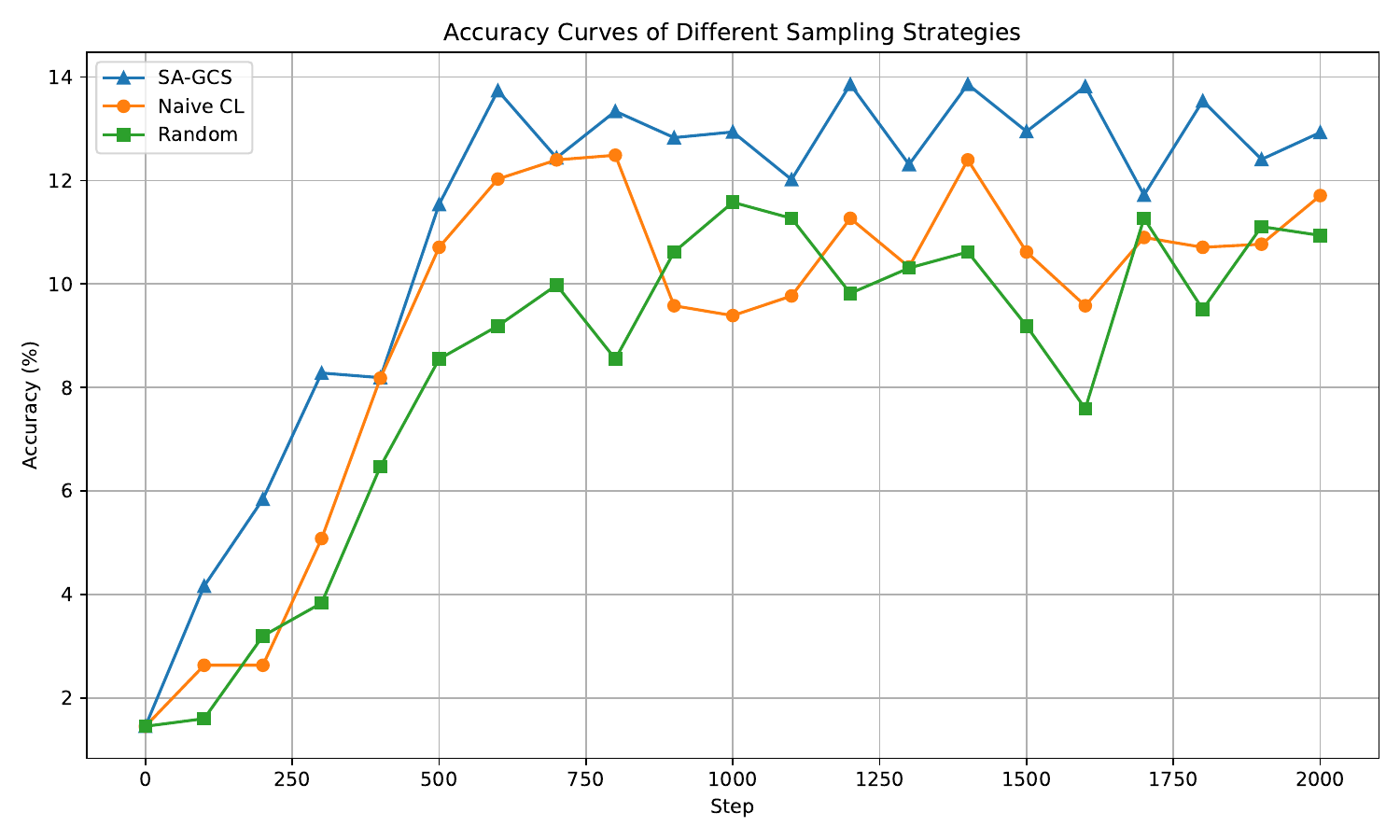} 
\caption{Success Rate progression on the Test-Unseen split using different sampling strategies (3B).}
\label{figure:step-sr-3b}
\end{figure}